\documentclass[letterpaper, 10 pt, conference]{ieeeconf}  

\IEEEoverridecommandlockouts                              

\usepackage{graphics} 
\usepackage{epsfig} 
\usepackage{mathptmx} 
\usepackage{times} 
\usepackage{amsmath} 
\usepackage{amssymb}  
\usepackage{graphicx}
\usepackage{cite}

\title{\LARGE \bf
This is the Way: Mitigating the Roll of an Autonomous Uncrewed Surface Vessel in Wavy Conditions Using Model Predictive Control*}

\author{Daniel Jenkins$^{1}$ and Joshua A.~Marshall$^{2}$
\thanks{*This research was funded in part by the NSERC Canadian Robotics Network (NCRN) under NSERC project NETGP 508451-17.}
\thanks{$^{1}$Daniel Jenkins is with the Ingenuity Labs Research Insitute and the Dept.\ of Electrical \& Computer Engineering,
        Queen's University, Kingston, ON K7L 3N6, Canada
        {\tt\small 23smd@queensu.ca}}%
\thanks{$^{2}$Joshua A.~Marshall is with the Ingenuity Labs Research Institute and the Dept.\ of Electrical \& Computer Engineering,
        Queen's University, Kingston, ON K7L 3N6, Canada
        {\tt\small joshua.marhsall@queensu.ca}}%
}

\begin{document}

\maketitle
\thispagestyle{empty}
\pagestyle{empty}


\begin{abstract}
Though larger vessels may be well-equipped to deal with wavy conditions, smaller vessels are often more susceptible to disturbances. This paper explores the development of a nonlinear model predictive control (NMPC) system for Uncrewed Surface Vessels (USVs) in wavy conditions to minimize average roll. The NMPC is based on a prediction method that uses information about the vessel's dynamics and an assumed wave model. This method is able to mitigate the roll of an under-actuated USV in a variety of conditions by adjusting the weights of the cost function. The results show a reduction of 39~\% of average roll with a tuned controller in conditions with 1.75-metre sinusoidal waves. A general and intuitive tuning strategy is established. This preliminary work is a proof of concept which sets the stage for the leveraging of wave prediction methodologies to perform planning and control in real time for USVs in real-world scenarios and field trials.

\end{abstract}

\section{INTRODUCTION}

A common challenge for some Uncrewed Surface Vessels (USVs) is operation in wavy conditions, which are complicated and difficult to predict \cite{liu}. Most USVs are under-actuated, and only able to provide direct control to the vessel's surge and yaw directions through linear and rotational forces provided by dual-propellers, or a propeller-rudder system. This results in an inability to mitigate disturbances in the roll and sway directions due to waves. These conditions result in issues including, but not limited to: decreased passenger comfort, loss of cargo, potential vessel damage, and decreased fidelity of control \cite{barrera}.

While existing research has focused on roll reduction with active elements such as rudder control \cite{Gallieri}, and improvements for USVs for dynamic positioning \cite{liu}, there is little literature on control and planning methods to reduce roll for active movement for under-actuated vessels. Controllers typically isolate low-frequency waves by filtering out individual wave action. However, this results in a loss of information that could prove important for optimal control. 

Nonlinear model predictive control (NMPC) has recently gained popularity in robotics due to its robustness in a variety of conditions, and the recent increase in computational capabilities that has previously been a limiting factor. An advantage of NMPC is also its ability to tackle problems not limited to direct control, such as simple planning tasks. 

\begin{figure}
    \centering
    \includegraphics[width=1\linewidth]{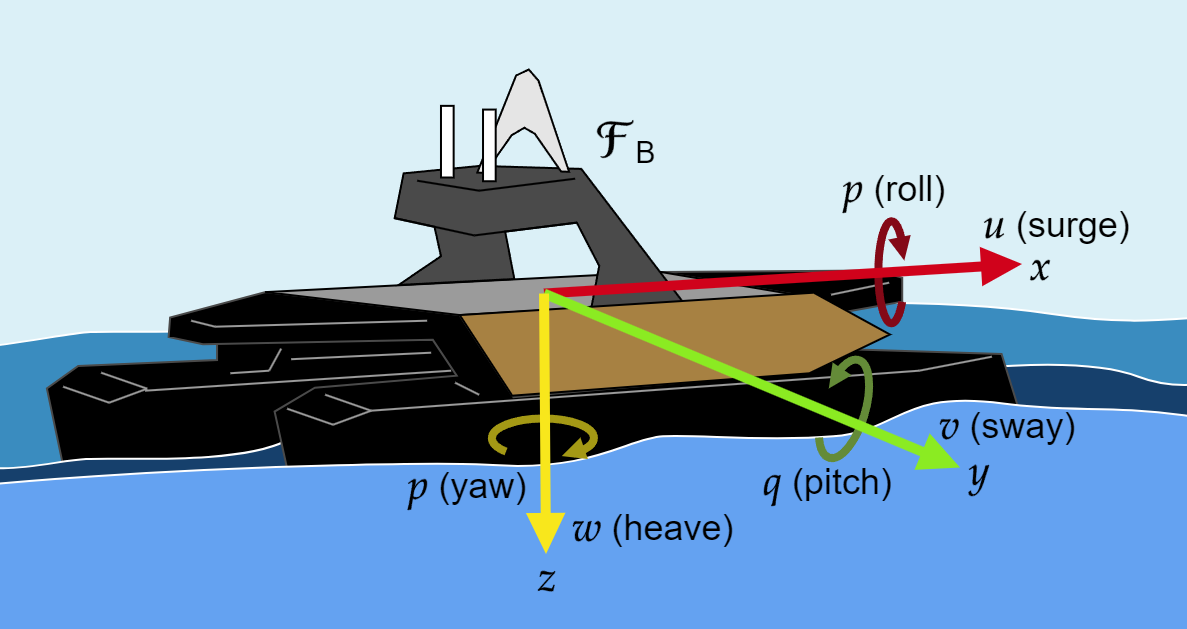}
    \caption{Positional and motion components of an uncrewed surface vessel (USV). Shown is a sketch of a Maritime Robotics Otter USV, the subject vessel of this paper. We use a North-East-Down (NED) body coordinate frame and the global positional frame is also NED.}
    \label{fig:ssyrph}
\end{figure}

This paper explores the development of a simulated NMPC controller that attempts to mitigate the roll of an under-actuated USV in wavy conditions.  In particular, we model a Maritime Robotics Otter USV, depicted schematically in Fig.~\ref{fig:ssyrph}, as the basis of the presented work. We study the emergence of unique vessel behaviours that compensate for wave-induced roll by adjusting the weights of the NMPC cost function, these behaviours supporting the potential efficacy of a similar controller under real-world circumstances. Finally, a systematic approach is presented for adjusting the weights based on desired performance for different sea states.

\section{RELATED WORK}
This section describes related work in the domains of NMPC, wave prediction, and dynamic positioning.

\subsection{MPC as Both Controller and Planner}
Though MPC has been used extensively in the chemical industry for decades, recent applications in robotics have demonstrated its potential to perform both control and planning. NMPC as a planner has been applied to autonomous driving at high speeds \cite{Sivas} and in urban settings \cite{Micheli},  as well as cooperative planning between Unmanned Aerial Vehicles (UAVs) and USVs, to safely land UAVs at sea, or perform ship-to-shore operations \cite{li}. Of particular note is the successful application of distributed NMPC for motion planning and coordination of multiple USVs \cite{Ferranti}. However, the current focus of NMPC for underactuated USVs is on trajectory tracking, which is not necessarily as vital in an open water scenario. Though some NMPC models utilize rudder roll control \cite{Qin}, this does not utilize the potential for NMPC to be used as both planner and controller.

\subsection{Wave Prediction}
A key challenge associated with operating in wavy conditions is to accurately predict the waves themselves. There has been recent progress in short-term wave prediction, particularly using data-driven algorithms such as neural networks and auto-regressive models, which are better able to capture the complicated nature of waves \cite{Ji,Deo}. This has proven effective in performing both short-term wave prediction and forecasts on buoys \cite{Ji}, wave-energy converters \cite{Fusco}, and stationary vessels \cite{Overaas}, as well as using similar techniques to improve models of vessel roll motion due to waves \cite{Zhang}. These methods predominantly utilize Inertial Measurement Units (IMUs) which are reliable in measuring the effect of waves on a vessel for a low cost.

One limiting factor is that the ability to accurately predict waves from a stationary vessel may not be capable of providing enough information for a control task like roll mitigation.  The work of Sears et al.~\cite{Sears} demonstrates a spatiotemporal mapping technique using Gaussian progress regression to predict the future wave conditions from previous measurements. Currently, this approach is limited to offline mapping, but a real-time variant would provide predictions of upcoming disturbances for the development of USV controllers in wavy conditions.

Although wave prediction is not part of this study, recent progress in this field motivates the discussion and research into designs for USV controllers that use real-time knowledge about wave conditions.
    
\subsection{Dynamic Positioning with NMPC}
Recently, to take advantage of developments in both wave prediction and NMPC, progress has been made in the design of NMPC controllers for use in Dynamic Positioning (DP) and Weather Optimal Control (WOC) of surface vessels \cite{Overaas,Halvorsen,Deng}. This has largely been driven by a desire to support underwater Remotely Operated Vehicles (ROVs) by using smaller vessels, which are more prone to movements from waves \cite{Overaas}, or to reduce overall energy consumption \cite{Deng}. 

The work of Øveraas et al.~\cite{Overaas} is of particular interest, in which a DP system is developed using auto-regressive wave prediction with NMPC in order to remove the delay associated with DP controllers. The wave prediction utilizes information about the vessel velocity and inputs to isolate environmental forces. However, this problem is solved with the use of an azimuth thruster that inherently removes some of the associated issues with under-actuated vessels. Nonetheless, their simulations provided promising results, showing that it is feasible to predict and act on short term wave action, significantly improving the DP capabilities of a vessel. In Halvorsen et al.~\cite{Halvorsen}, a similar setup is considered with increased variation. Of interest is the development of a roll minimizing controller using velocity feedback, which successfully decreases roll using an azimuth thruster. 

\section{PROBLEM FORMULATION}
The problem setup and simulation results presented in this section are based on the Maritime Robotics Otter USV\footnote{See https://www.maritimerobotics.com/otter.}.

\subsection{Vessel Kinematics and Dynamics}
For the purposes of vessel control, it is common practice to simplify the model by using only three degrees of freedom (DOF). However, this is insufficient for the purposes of this study, in which roll dynamics are required. For our purposes, a full 6-DOF model is required. The vessel's state for position ($\boldsymbol{\eta}$), velocity ($\boldsymbol{\nu}$), and forces and moments ($\boldsymbol{\tau}$) are given by \cite[Section 2.1]{Fossen}
\begin{equation}
\begin{split}
    \boldsymbol{\eta}
&= \begin{bmatrix}
        x & 
        y & 
        z & 
        \phi & 
        \theta &
        \psi
    \end{bmatrix}^\top \\
    \boldsymbol{\nu} &= 
    \begin{bmatrix}
        u &
        v &
        w &
        p & 
        q &
        r 
    \end{bmatrix}^\top \\
    \boldsymbol{\tau} &= \begin{bmatrix}
        X & Y & Z & K & M & N
    \end{bmatrix}^\top, 
\end{split}
\end{equation}
corresponding to the associated $x$, $y$, $z$, roll ($\phi$), pitch ($\theta$), and yaw ($\psi$) axes and rotations respectively, for each vector. With $\boldsymbol{\eta}$ in reference to a North-East-Down (NED) global frame, and $\boldsymbol{\nu}$ and $\boldsymbol{\tau}$ with reference to the body frame. The derivative  $\boldsymbol{\dot{\eta}}$ and the respective change in coordinate frame is related to $\boldsymbol{\nu}$ via
\begin{equation}
    \dot{\boldsymbol{\eta}} = \boldsymbol{R} (\phi,\theta,\psi) \nu.
\end{equation}
The vessel dynamics are of the form \cite{Fossen},
\begin{equation}
    \mathbf{M}\boldsymbol{\dot{\nu}}+\mathbf{C}(\boldsymbol{\nu})\boldsymbol{\nu}+\mathbf{D}(\boldsymbol{\nu})\boldsymbol{\nu} + \mathbf{G}(\eta) = \mathbf{U} + \boldsymbol{\tau}_{\rm wave}
\end{equation}
where
\begin{equation}
\mathbf{U} = \begin{bmatrix}\tau_X & 0 & \dots & 0 & \tau_N \end{bmatrix}^\top,
\end{equation}
$\mathbf{M}$ is the system's mass matrix, $\mathbf{C}$ is the Coriolis and centripetal matrix, $\mathbf{D}$ is the damping matrix, and $\mathbf{G}$ captures gravity and buoyancy forces and moments.
The matrices $\mathbf{M}$ and $\mathbf{C}$ can also be broken into the rigid-body and added hydrodynamic portions
\begin{equation}
\begin{split}
    \mathbf{M} &= \mathbf{M}_{RB} + \mathbf{M}_A \\
    \mathbf{C} &= \mathbf{C}_{RB} + \mathbf{C}_A,
\end{split}
\end{equation} respectively. Due to the complexity of hydrodynamics, the specific values of the associated matrices need to estimated empirically. For the Otter USV modelled in this paper, these empirical values were obtained from Fossen~\cite{Fossen} and associated code repository\footnote{https://github.com/cybergalactic/PythonVehicleSimulator}. We assume that the added mass terms do not vary significantly from their steady-state/null frequency values.

\subsection{Wave Model and Simulation}
Realistic waves can be best modelled with a series of sinusoids or time-series data of real wave recordings \cite{Frechot}. For this study, our emphasis is on the response to a dominant sinusoidal portion of the waves. We model the waves directly as a single sinusoid because the focus of this work is not a robust study of the vessels dynamic response to real waves, but rather the behaviour of the NMPC when presented with ocean wave-like phenomena. This can be described as a function for the elevation of the wave at a point and time 
\begin{equation}
    f(y,t) = \frac{H_w}{2}\left(\frac{2\pi}{\lambda}y +  \frac{2\pi}{T_w}t\right)
\end{equation}
where $H_w$, $\lambda$, and $T_w$ represent the wave's height (m), wavelength (m), and period (s), respectively. Empirical wave data is then used to create an ``average'' wave system; see Table \ref{tab:wavestates}. The derived gradient
\begin{equation}
    \nabla_y f(y,t) = H_w \frac{\pi}{\lambda}\cos\left(\frac{2\pi}{\lambda}y +  \frac{2\pi}{T_w}t\right)
\end{equation}
allows the a description of the wave angle $\alpha$ as
\begin{equation}
    \alpha = \arctan\left(\nabla_y f(y,t)\right).
\end{equation} 

\begin{table}
\end{table}

\begin{table}
\caption{NATO Annual Sea States in the Open Ocean, Northern Hemisphere \cite{Huston}}
\label{tab:wavestates}
\resizebox{\columnwidth}{!}{%
\begin{tabular}{|c|c|c|c|c|}
\hline
{Sea-State} &
  {\begin{tabular}[c]{@{}c@{}}Significant\\ Wave Height\\ (m)\end{tabular}} &
  {\begin{tabular}[c]{@{}c@{}}Sustained\\ Wind Speed\\ (kts)\end{tabular}} &
  {\begin{tabular}[c]{@{}c@{}}Modal Wave\\ Period Range\\ (s)\end{tabular}} &
  {\begin{tabular}[c]{@{}c@{}}Modal Wave\\ Period Most \\ Probable\\ (s)\end{tabular}} \\ \hline\hline
0-1             & 0 - 0.1           & 0 - 6           & -        & -  \\ \hline
2               & 0.1 - 0.5         & 7 - 10          & 3 - 15   & 7  \\ \hline
3               & 0.5 - 1.25        & 11 - 16         & 5 - 15.5 & 8  \\ \hline
4               & 1.25 - 2.5        & 17 - 21         & 6 - 16   & 9  \\ \hline
5               & 2.5 - 4.0         & 22 - 27         & 7 - 16.5 & 10 \\ \hline
6               & 4.0 - 6.0         & 28 - 47         & 9 - 17   & 12 \\ \hline
7               & 6.0 - 9.0         & 48 -55          & 10 - 18  & 14 \\ \hline
8               & 9.0 - 14.0        & 56 - 63         & 13 - 19  & 17 \\ \hline
\textgreater{}8 & \textgreater 14.0 & \textgreater 63 & 18 - 24  & 20 \\ \hline
\end{tabular}%
}
\end{table}

\subsection{Vessel's Response to Waves}
A vessel's response to waves requires a numerical analysis of the vessel's geometry via Response Amplitude Operators (RAOs), a series of statistics regarding the specific response of a vessel to given wave parameters, and can be obtained from tools such as WAMIT or MARIN \cite[Section 10.2]{Fossen}. However, for the purposes of control system design, this can be simplified. The key objective for the simulation in this study is to create motions similar to what is expected rather than a complete wave simulation. The rotational forces are approximated using the vessels dynamics on a flat water surface. Through manipulation of the inputs of the $\mathbf{G}(\eta)$, \begin{equation}\mathbf{M}\boldsymbol{\dot{\nu}}+\mathbf{C}(\boldsymbol{\nu})\boldsymbol{\nu}+\mathbf{D}(\boldsymbol{\nu})\boldsymbol{\nu} + \mathbf{G}(\beta) = \mathbf{U} + \boldsymbol{\tau}_{\rm wave},
\end{equation}
where
\begin{equation}
    \beta = \begin{bmatrix}
        x & 
        y & 
        z & 
        \phi - \alpha\cos(\psi) & 
        \theta - \alpha\sin(\psi)  & 
        \psi
    \end{bmatrix}^\top,
\end{equation} 
the roll and pitch terms instead consist of the relative angle to the surface of the water rather than to the $x$-$y$ plane, and the vessel reacts to restore its position parallel to the water's surface along the wave.

\begin{figure}
    \centering
    \includegraphics[width=0.75\linewidth]{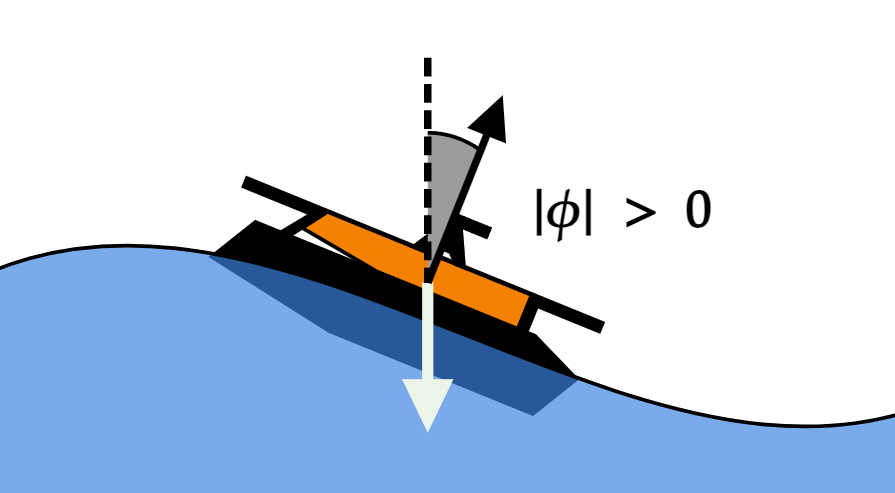}
    \caption{Gravitational and buoyancy force vectors for a tilted vessel with pitch. The lateral force can be extrapolated from the buoyancy component, whose magnitude is dependent on gravity. Figure adapted from \cite{Sears}.}
    \label{fig:buoy}
\end{figure}

\begin{figure}
    \centering
    \includegraphics[width=1\linewidth]{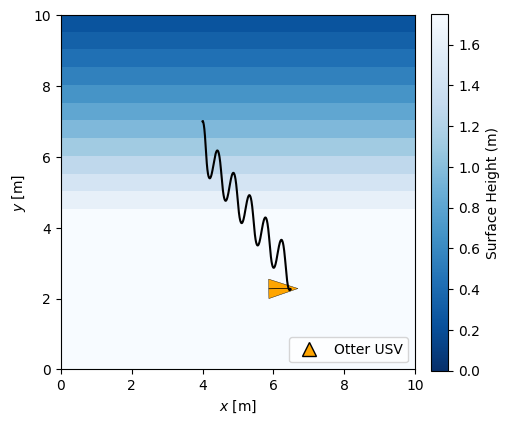}
    \caption{Given a small forward thrust, the vessel experiences a general trend in the direction of the waves due to the underlying current. }
    \label{fig:wave-motion}
\end{figure}

The lateral portion was estimated by isolating the instantaneous slope of the water and applying an equal buoyancy force to that of gravity (see Fig. \ref{fig:buoy})
\begin{equation}
    F_{w} = \frac{mg}{2}\sin(2\alpha),
\end{equation}
resulting in an oscillatory movement in the direction of the wave motion (see Fig. \ref{fig:wave-motion}).

\subsection{NMPC Design}
With a model established for the NMPC, the behaviour of the controller can be then designed through the use of a cost function $J$ to be minimized,
\begin{equation}
\label{eqn:costmin}
    \min_{\eta,\mathbf{U}} J(\eta,\mathbf{U},k)
\end{equation}
\[
\begin{split}
    \text{s.t.} \quad \boldsymbol{\eta}_{k+1} = \boldsymbol{\eta}_k + \boldsymbol{\dot{\eta}}T \quad \forall k \in [0,N] \\
    \boldsymbol{\nu}_{k+1} = \boldsymbol{\nu}_k + \boldsymbol{\dot{\nu}}T \quad \forall k \in [0,N]\\
    \begin{bmatrix} -\tau_{X,\max} \\ -\tau_{N,\max} \end{bmatrix}\leq \boldsymbol{U_k} \leq \begin{bmatrix} \tau_{X,\max} \\ \tau_{N,\max} \end{bmatrix} \quad \forall k \in [0,N] \\
    \boldsymbol{\eta}_0 = \begin{bmatrix}
        0 & \dots & 0 & \frac{\pi}{2}
    \end{bmatrix}^T\textup{~and~}
    \boldsymbol{\nu}_0 = \mathbf{0}
\end{split}
\]
where $T>0$ is the sampling period (s), $k\in\mathbb{N}^+$ is the current timestep, and $N>0$ is the total number of timesteps. The cost function can be thought of as encouraging the vessel to reach a waypoint in open ocean while also minimizing roll throughout its travel. 

Reaching the waypoint can be achieved with various strategies. One is to assign the vessel a minimum speed, ($u_k \geq 0.5 ~ \forall k$) and have a cost for the heading error
\begin{equation}
    e(\boldsymbol{\eta})_\psi = \arctan\left(\frac{y_d-y}{x_d-x}\right) - \psi.
\end{equation}
This can be combined with a second, distance, cost
\begin{equation}
    d(\boldsymbol{\eta}) = \sqrt{(y_d-y_k)^2+(x_d-x_k)^2}
\end{equation}
which rewards progress and therefore speed to the goal.

Most relevantly, we added a cost to minimize the roll of the vessel ($\phi$). To prevent oscillations of the inputs, an input derivative term was added. The resultant cost function $J$ for a given prediction horizon $P$ is then
\begin{equation}
    J(\boldsymbol{\eta},\mathbf{U},k) = \sum_{i=k}^{k+P} 
    Qe_{\psi,i}^2 + 
    R\phi_i^2 + 
    Sd_i + \mathbf{\dot{U}}_i^T\mathbf{W}\mathbf{\dot{U}}_i
\end{equation}
where $Q$, $R$, $S$, $\mathbf{W}$ are the respective cost weights. NMPC was implemented using the above models and cost function in Python and with the CasADi IPOPT solver for optimization.

\section{RESULTS \& DISCUSSION}

To study the operation of the NMPC controller, a series of simulations were performed to illustrate the effects each cost has on the overall behaviour. The vessel was simulated in a Sea State of 4 \cite{Huston} (Table \ref{tab:wavestates}, Table \ref{tab:parameters}).

\begin{table}
\end{table}
\begin{table}
\caption{Simulation Parameters}
\label{tab:parameters}
\resizebox{\columnwidth}{!}{%
\begin{tabular}{|ccc|cc|cc|}
\hline
\multicolumn{3}{|c|}{Wave}                                                       & \multicolumn{2}{c|}{NMPC}               & \multicolumn{2}{c|}{Simulation}            \\ \hline\hline
\multicolumn{1}{|c|}{$T_w$ (s)} & \multicolumn{1}{c|}{$\lambda$ (m)} & $H_w$ (m) & \multicolumn{1}{c|}{$\mathbf{W}$} & $P$ & \multicolumn{1}{c|}{$T$ (s)} & $t_{\max}$ (s) \\ \hline
\multicolumn{1}{|c|}{6.0}       & \multicolumn{1}{c|}{35.0}            & 1.75      & \multicolumn{1}{c|}{diag(1,1)}    & 40  & \multicolumn{1}{c|}{0.1}     & 180         \\ \hline
\end{tabular}%
}
\end{table}
\begin{table}
\caption{Roll and Time results for Various NMPC Weighting Combinations}
\label{tab:results}
\resizebox{\columnwidth}{!}{%
\begin{tabular}{|l|c|c|c|c|}
\hline
\multicolumn{1}{|c|}{Description} & \begin{tabular}[c]{@{}c@{}}Weights \\ ($Q,R,S$)\end{tabular} & \multicolumn{1}{c|}{\begin{tabular}[c]{@{}c@{}}Avg. Roll \\ ($\mu_\phi$ {[}deg{]})\end{tabular}} & \multicolumn{1}{c|}{\begin{tabular}[c]{@{}c@{}}Max. Roll \\ ($\phi_{\max}$ {[}deg{]})\end{tabular}} & \multicolumn{1}{c|}{\begin{tabular}[c]{@{}c@{}}Time to Wpt. \\ (s)\end{tabular}} \\ \hline\hline
Direct & 5, 0, 2 & 4.00 & 7.64 & 59.7 \\ \hline
Indirect & 5, 750, 2 & 3.65 & 9.10 & 65.9 \\ \hline
Low Roll & 5, 2500, 2 & 1.79 & 8.10 & N/A \\ \hline
Low $Q$ & 2, 1550, 2 & 1.45 & 2.77 & N/A \\ \hline
Low Tack & 5, 1000, 2 & 3.25 & 9.27 & 73.6 \\ \hline
Balanced & 5, 1550, 2 & 2.46 & 8.82 & 97.3 \\ \hline
\end{tabular}%
}
\end{table}

This Sea State is representative of ``moderate waves", which would be normally troublesome for a small vessel such as the Otter USV. The simulation was terminated if the vessel reached within 2 m of the waypoint $(x_d,y_d)=(85,75)$.  A series of runs was performed, with varying weights on the cost function to influence and observe different behaviours. The prediction horizon of 40 time steps was selected for the controller to exhibit planning behaviour, as larger horizons were computationally expensive for little added benefit. The results are shown in Table \ref{tab:results}, with each iteration taking an average of 0.05 seconds to compute, indicating real-time applicability.

\begin{figure}
    \centering
    \includegraphics[width=0.9\linewidth]{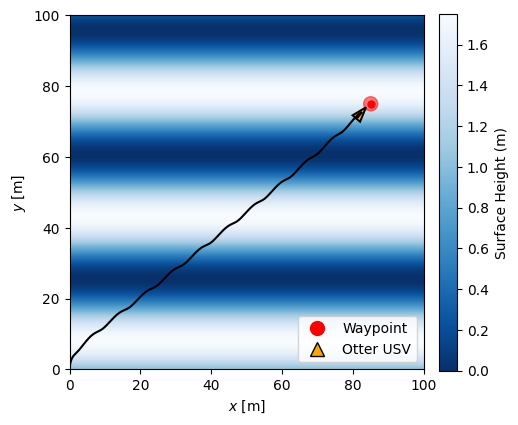}
    \caption{The path taken by the ``direct" controller. With no roll cost, the vessel experiences small variations due to the wave forces, but generally acts on a direct path to the waypoint.}
    \label{fig:direct_path}
\end{figure}

\begin{figure}
    \centering
    \includegraphics[width=0.9\linewidth]{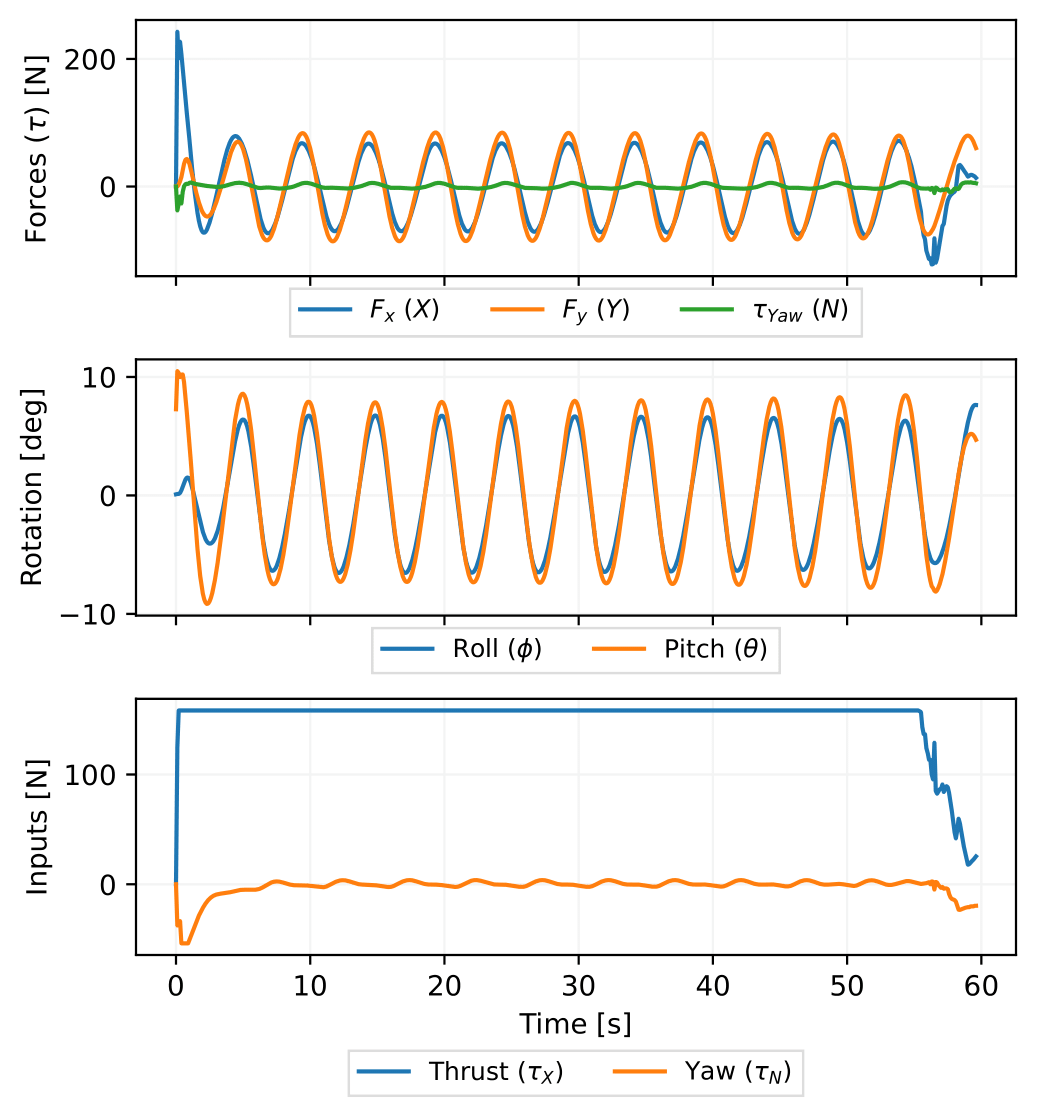}
    \caption{Forces, roll, pitch, and inputs for ``direct" controller. Pitch and Roll remain relatively equal throughout the path. The vessel is not disturbed largely by these waves, by experiences a significant amount of roll due to its heading with respect to the waves.}
    \label{fig:direct_graph}
\end{figure}

A standard ``Direct" route did not consider the roll of the vessel and instead optimized a direct path to the waypoint with no roll cost. Fig.~\ref{fig:direct_path} shows the path taken by the USV in this configuration. This serves as a baseline for a typical traversal when not prioritizing roll (see Fig.~\ref{fig:direct_graph}). 

\begin{figure}
    \centering
    \includegraphics[width=0.9\linewidth]{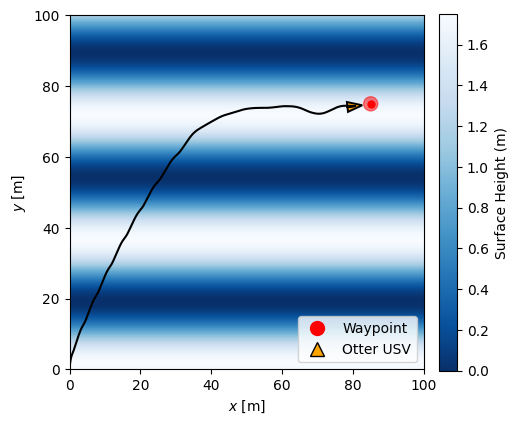}
    \caption{The path taken by ``indirect" controller. The vessel initially heads more into the direction of the waves, but as it approaches the waypoint it abandons this strategy and moves directly towards it.}
    \label{fig:indirect_path}
\end{figure}

\begin{figure}
    \centering
    \includegraphics[width=0.9\linewidth]{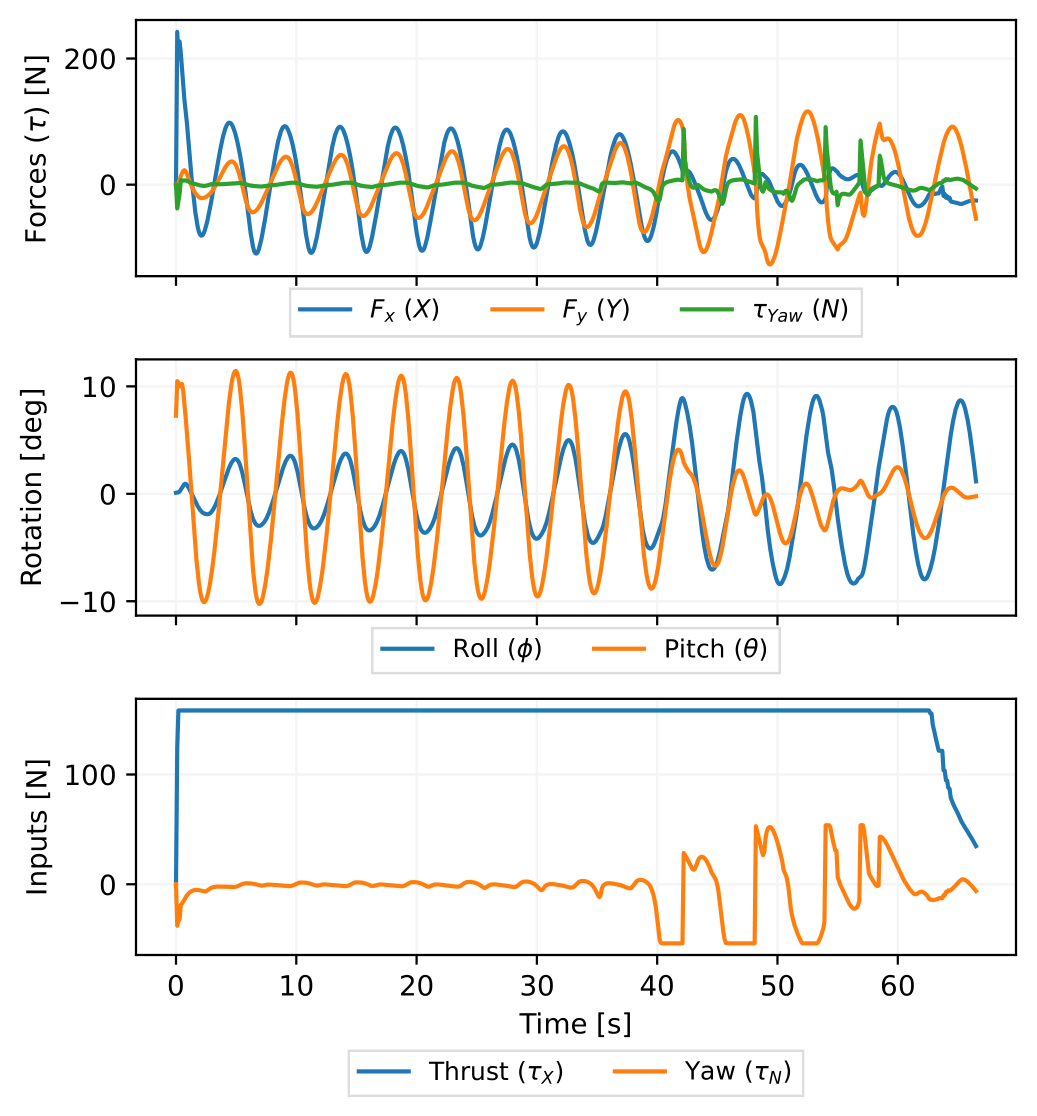}
    \caption{Forces, roll, pitch, and inputs for ``indirect" controller. The vessel initially experiences lower roll than the ``direct'' path, but begins to move parallel to the waves and experiences larger roll to reach the waypoint. }
    \label{fig:indirect_graph}
\end{figure}

Small levels of roll cost results in a relatively minor overall reduction in average roll (Fig.~\ref{fig:indirect_path}). However, the roll is distributed much less evenly, and now the vessel experiences large amounts of roll closer to the waypoint, as the heading error begins to dominate the roll term that previously dictated motion (Fig.~\ref{fig:indirect_graph}). 

A large roll cost has negative effects because the vessel no longer prioritizes reaching the waypoint. Rather, the vessel prioritizes navigating perpendicular to the waves, without waypoint costs becoming high enough to overcome the roll cost. This could be beneficial when it may be advantageous to wait for a calming of the waves to perform riskier high-roll turns. A similar path was observed if the heading error cost remains too small. This is because the controller is never incentivized to turn around or tack.

\begin{figure}
    \centering
    \includegraphics[width=0.9\linewidth]{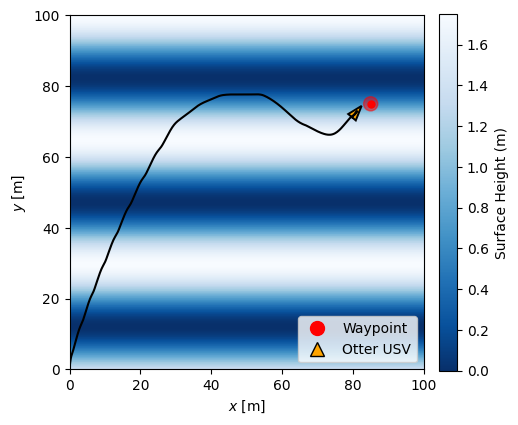}
    \caption{The path taken by the ``low tack" controller. Initially the vessel moves similarly to the ``indirect'' controller. Instead of moving directly towards the waypoint, it tacks into the waves whilst moving horizontally. }
    \label{fig:lowtack_path}
\end{figure}

\begin{figure}
    \centering
    \includegraphics[width=0.9\linewidth]{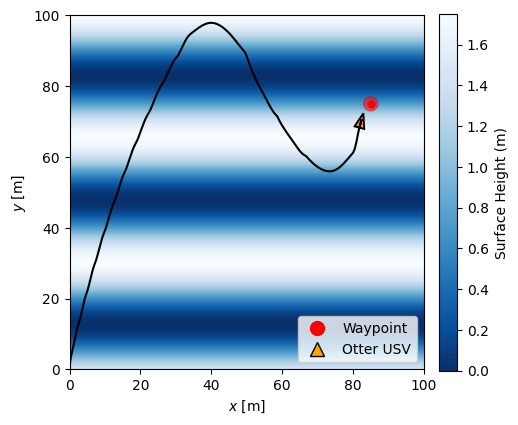}
    \caption{The path taken by the ``balanced" controller. The vessel begins to show tacking behaviour. In order to minimize roll, it oscillates its path with respect to the $y$-axis, whilst achieving progress along the $x$-axis.}
    \label{fig:balanced_path}
\end{figure}

At a more balanced level of roll cost, the vessel begins to display behaviour similar to that of a human operator, tacking (Fig.~\ref{fig:lowtack_path}). Increasing this roll component further encourages this behaviour (Fig.~\ref{fig:balanced_path}). Overall, an effective 39~\% reduction in average roll can be seen compared to the direct case. This however, is at the cost of increased time to reach the waypoint, and an increased maximal roll of 15~\%. The direct controller has the lowest maximum roll of all the successful controllers because it does not turn around and experience momentary high roll. It should also be noted that this control strategy inherently minimizes the time where waves are abeam to the vessel, which is typically the cause for capsizing in small vessel tests \cite{AlSalem}.

\begin{figure}
    \centering
    \includegraphics[width=0.9\linewidth]{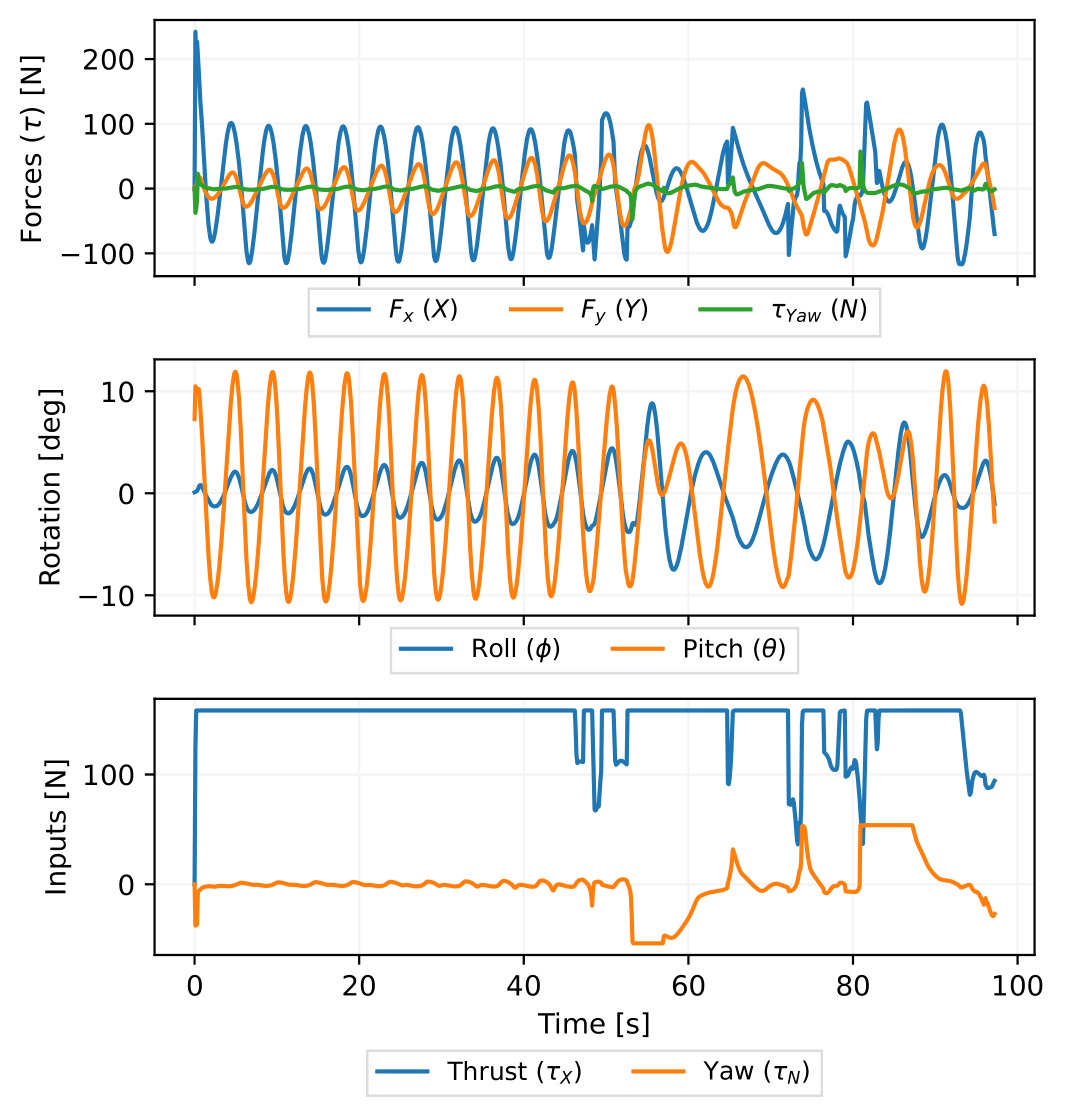}
    \caption{Forces, pitch and roll, and inputs for ``balanced" controller. On average, the roll is lower than other cases. However, during the turn (50~s), the vessel experienced higher roll since the waves have not calmed. Average roll is reduced and the vessel experiences minimal disturbance.}
    \label{fig:balanced_graph}
\end{figure}

The emergence of tacking behaviour is promising. It shows capabilities for an NMPC controller to act as both a controller for maintaining waypoint navigation, but also a planning tool to avoid large amounts of roll. When the controller is given proper constraints on its scenario, it reacts with intuitive behaviour which is often absent in typical controllers. If similar behaviour is desired across different intensities of Sea States, re-tuning of the controller is required since inherently rougher seas will always result in higher roll. However, this exploration has led to a clear and very intuitive method of tuning this type of controller to achieve desired behaviour, though currently through some trial and error. The steps to do so are to:
\begin{enumerate}
    \item Tune $Q$ and $S$ to obtain a good direct path model
    \item Increase $R$ until desired behaviour is reached or vessel no longer aims for waypoint
    \item If vessel ceases to aim for waypoint, increase $Q$
    \item If vessel makes slow/poor progress to waypoint, increase $S$
\end{enumerate}
However, a controller can be tuned a single time for a given vessel to avoid roll of certain magnitude in rougher sea states in order to wait for a calmer sea state.

\section{CONCLUSION}
Work has begun on the development of an NMPC controller for USV navigation in wavy conditions to minimize the average roll. This is achieved through the introduction of a roll cost. This work shows an reduction of 39\% of average roll with a tuned controller in conditions with 1.75-metre sinusoidal waves The resulting motions of this controller are similar to those expected of a human operator. This preliminary work is a proof of concept which sets the stage for the leveraging of wave prediction methodologies to perform prediction and control in real time for USVs in real-world scenarios and field trials.

The next step for this work is to integrate it with a wave prediction model instead of feeding it ground truth. The system can then be simulated in increasingly complicated wave scenarios before testing the system on a physical Otter USV in real-world trials. 

\bibliographystyle{IEEEtran}
\bibliography{IEEEabrv,root}

\end{document}